\documentclass{article} % For LaTeX2e
\usepackage{iclr2026_conference,times}

\usepackage{amsmath,amsfonts,bm}

\def\eqref#1{equation~\ref{#1}}
\def\1{\bm{1}}

\DeclareMathAlphabet{\mathsfit}{\encodingdefault}{\sfdefault}{m}{sl}
\SetMathAlphabet{\mathsfit}{bold}{\encodingdefault}{\sfdefault}{bx}{n}

\usepackage{wrapfig}
\usepackage{hyperref}
\usepackage{url}
\usepackage{tikz}
\usepackage{multirow}
\usepackage{booktabs}
\usepackage{enumitem}
\usepackage{xspace}
\usepackage{adjustbox}
\usepackage{tabularray}
\usepackage{booktabs,tabularx}
\usepackage{wrapfig}
\usepackage{verbatim}
\usepackage{fancyvrb}
\usepackage{fvextra}
\usepackage{scalerel}
\usepackage{pifont}
\usepackage{subcaption}
\usepackage[cachedir=minted-cache]{minted}
\usepackage{amsmath, amssymb} 
\usepackage{adjustbox}        
\usepackage{tcolorbox}        
\usepackage{listings}

\usepackage{color}
\definecolor{keywordcolor}{rgb}{0.7, 0.1, 0.1}   % red
\definecolor{tacticcolor}{rgb}{0.0, 0.1, 0.6}    % blue
\definecolor{commentcolor}{rgb}{0.4, 0.4, 0.4}   % grey
\definecolor{symbolcolor}{rgb}{0.0, 0.1, 0.6}    % blue
\definecolor{sortcolor}{rgb}{0.1, 0.5, 0.1}      % green
\definecolor{attributecolor}{rgb}{0.7, 0.1, 0.1} % red
\title{Lean4Physics: Comprehensive Reasoning Framework for College-level Physics in Lean4}
\newcommand{\method}{\textsf{{Lean4PHYS}}\xspace}
\newcommand{\bench}{\textit{LeanPhysBench}\xspace}
\newcommand{\lib}{\textit{PhysLib}\xspace}

\author{
    \textbf{Yuxin Li\textsuperscript{1}}\thanks{First Authors}, 
    \textbf{Minghao Liu\textsuperscript{1}}\footnotemark[1], 
    \textbf{Ruida Wang\textsuperscript{2}}\footnotemark[1],
    \textbf{Wenzhao Ji\textsuperscript{1}}, 
    \textbf{Zhitao He\textsuperscript{1}}, 
    \textbf{Rui Pan\textsuperscript{2}}, \\
    \textbf{Junming Huang\textsuperscript{3}}, 
    \textbf{Tong Zhang\textsuperscript{2}}, 
    \textbf{Yi R. (May) Fung\textsuperscript{1}}
    \\  
        \textsuperscript{1}Hong Kong University of Science and Technology, \\
        \textsuperscript{2}University of Illinois Urbana-Champaign, \\
        \textsuperscript{3}Princeton University
    \\
        \texttt{\{ylinq,mliuby,wjiab,zhebu\}@connect.ust.hk} \\
        \texttt{\{ruidaw,ruip4\}@illinois.edu} \\
        \texttt{junminghuang@princeton.edu} \\
        \texttt{tongzhang@tongzhang-ml.org} \\
        \texttt{yrfung@ust.hk} 
}

\iclrfinalcopy
\begin{document}

\maketitle

\begin{abstract}
    We present \method, a comprehensive reasoning framework for college-level physics problems in Lean4. 
    \method includes \bench, a college-level benchmark for formal physics reasoning in Lean4, which contains 200 hand-crafted and peer-reviewed statements derived from university textbooks and physics competition problems. 
    To establish a solid foundation for formal reasoning in physics, we also introduce \lib, a community-driven repository containing fundamental unit systems and theorems essential for formal physics reasoning. 
    Based on the benchmark and Lean4 repository we composed in \method, we report baseline results using major expert Math Lean4 provers and state-of-the-art closed-source models, with the best performance of DeepSeek-Prover-V2-7B achieving only \textbf{16\%} and Claude-Sonnet-4 achieving \textbf{35\%}. 
    We also conduct a detailed analysis showing that our \lib can achieve an average improvement of \textbf{11.75\%} in model performance. 
    This demonstrates the challenging nature of our \bench and the effectiveness of \lib. 
    To the best of our knowledge, this is the first study to provide a physics benchmark in Lean4.\footnote{Our code and dataset will be released at \url{https://github.com/ShirleyLIYuxin/Lean4PHYS}.} 
    
    % \red{TODO: Create a new publicly available github repo for that, we don't want to risk the exposore of our process of development}
    
    % We also analysis their performance qualitatively and quantitatively. Most of the expert provers do not outperform general models as they did in the math domain. This indicates a potential overfitting of math domain for current provers
    
    % , and provide an analysis of their performance. In the experiment, we identify that most expert provers do not outperform general models as they did in the math domain. This indicates a potential overfitting of the math domain rather than learning formal reasoning. We also conduct a comprehensive experiment showing that with \lib in the context, LLMs' performance on \bench increases by \textbf{11.88\%} on average, proving the effectiveness of our repository in assisting LLMs to solve the Lean4 physics problem. 
    % To the best of our knowledge, we are the first study to provide a physics benchmark in Lean4.\footnote{Our code and dataset will be released at \url{https://github.com/ShirleyLIYuxin/Lean4PHYS.git}.}
\end{abstract}

\section{Introduction}\label{sec:intro}

Formal thinking capability has long been considered a cornerstone of human intelligence and a key objective of machine learning. With the emergence of Large Language Models (LLMs), many studies explore diverse ways to apply LLMs to perform various reasoning tasks. These efforts span general reasoning~\citep{wang2024mmlu, suzgun2022challenging, talmor2018commonsenseqa}, mathematical reasoning~\citep{hendrycks2021measuring, cobbe2021gsm8k, guo2025deepseek}, natural science reasoning~\citep{saikh2022scienceqa, edwards2025mclm}, and various other domains~\citep{he2025verir1precisefaithfulclaim,su2025thinkingimagesmultimodalreasoning}. However, most works treat reasoning as a purely Natural Language (NL) task, relying on answer checking to evaluate correctness. This approach often fails to verify the intermediate steps of the reasoning process.

To make the reasoning process verifiable, researchers ground the reasoning procedure in formal logical systems, enabling the automatic verification of both the reasoning steps and final results through Formal Languages (FLs). Based on this approach, several FLs are developed, including Lean~\citep{de2015lean, moura2021lean}, Coq~\citep{coq1996coq}, Isabelle~\citep{paulson1994isabelle}, and HOL~\citep{harrison2009hol}. Among these, Lean4 has received significant attention from both academia and industry, making it one of the most well-studied FLs in recent years. Numerous works have been proposed to advance Lean4 research, including test benchmarks~\citep{zheng2021minif2f, gulati2024putnam, azerbayev2023proofnet}, training dataset~\citep{dong2025stp, wang2024theoremllama, ying2024lean}, and provers~\citep{polu2022formal, wang2024theoremllama, xin2024deepseek, wang2025ma, lin2025goedel2, dong2025stp, xin2025bfs, ren2025deepseek}. 

\begin{figure}[t]
    \centering
    \includegraphics[width=0.98\textwidth]{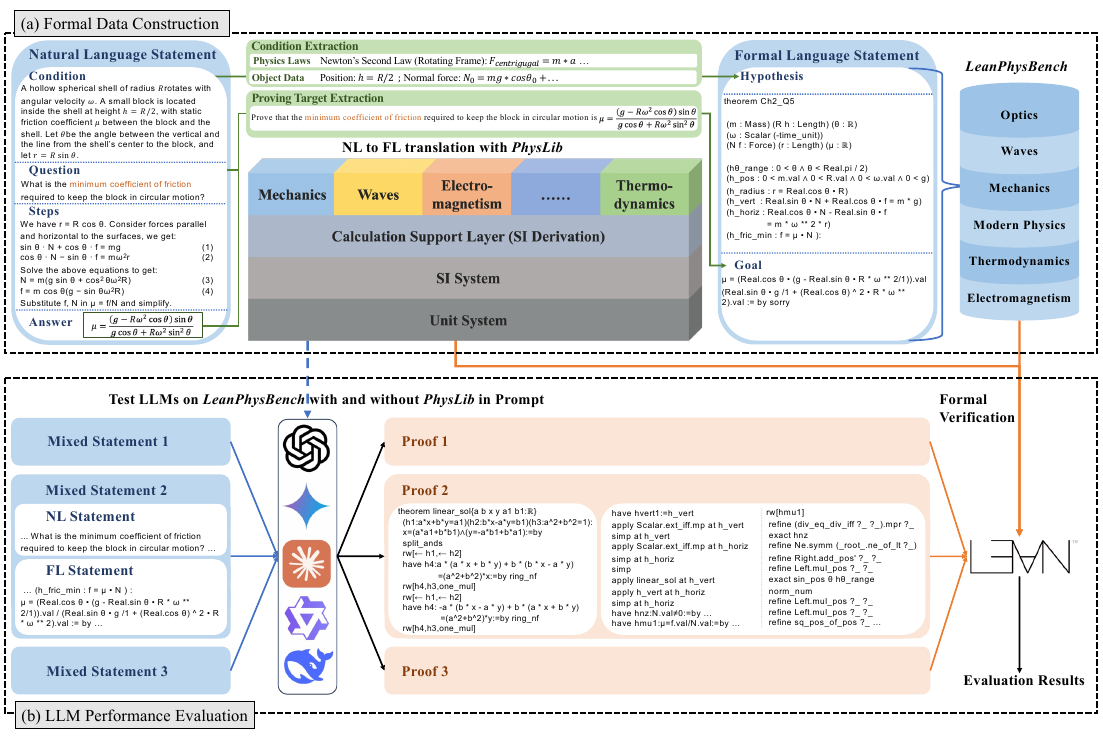} 
    \caption{
        \textbf{Overview of the \method framework:} (a) Benchmark and Library Construction: The benchmark and library are developed using a bottom-up principle. We first establish the foundational units and SI system of \lib, then compose the calculation support, and finally construct the field-related theorems. For the benchmark, NL questions are transformed from a question-answering format to proof problems and subsequently formalized into Lean4 statements. (b) LLM Performance Evaluation: We evaluate the formal physics reasoning capabilities of major open and closed-source LLMs on \bench, both with and without \lib. After inference is completed, Lean4 auto-verification evaluates model performance and highlights the utility of \lib.
        % it consists of two components for supporting LLM formal physics reasoning. 
        % (1) \lib: an extensible repository providing a physics unit system and commonly used theorems. 
        % (2) \bench: a benchmark of 200 hand-crafted theorems from high school competitions to elementary college level, designed to evaluate LLMs’ Lean4 physics reasoning.
    }
    \label{fig:Lean4PHYS_overview}
    \vspace{-0.2in}
\end{figure}

However, current studies in formal reasoning primarily focus on the math domain, leaving other fields with solid foundations that can be formalized, such as physics, largely understudied. This raises two critical challenges for Lean4 physics reasoning. Firstly, there is a lack of a comprehensive foundational repository to support physics formalization. Unlike mathematics, which benefits from extensive libraries like Mathlib, physics reasoning requires specialized infrastructure, such as unit systems and domain-specific theorems. Such a foundation is absent or fragmented. Secondly, there is no thorough evaluation of the models' capabilities in formal physics reasoning. While state-of-the-art expert provers demonstrate their superior performance on Lean4 Math benchmarks such as MiniF2F~\citep{zheng2021minif2f} and PutnamBench~\citep{gulati2024putnam}, it remains unclear whether their formal reasoning capabilities also apply to physics problems. Or these models have overfit to math reasoning patterns despite using similar Lean4 syntax.

To solve the problem above, we propose \method, a comprehensive reasoning framework for college-level general-domain physics problems in Lean4. It aims to provide a foundation for LLM-based formal physics reasoning. \method launches \lib, a foundation repository that supports Lean4 physics reasoning. The \lib builds upon the dedicated \textit{UnitSystem} in \citet{teorth_analysis}. We implement the basic mathematical structure, calculus, and algebraic operations, transfer between physics and math units, dimension transformation, and a growing collection of extendable formalized physics theorems. To provide a baseline verifying LLMs' formal reasoning capability, we develop \bench based on \lib that contains 200 manually crafted and peer-reviewed Lean4 theorems. \bench targets college-level knowledge, spanning Olympiad-style competition problems and college textbook-level physics problems formalized in Lean4. Such a benchmark is composed by systematically transforming natural-language problems into Lean4 theorems: extracting key conditions and relevant physical laws, defining explicit proving targets, restating the problems in logical form, and integrating them into verifiable Lean4 statements. To the best of our knowledge, \bench is the first benchmark to evaluate LLMs' formal physics reasoning capabilities. 

We summarize our contributions as follows:
\begin{enumerate}
    \item We introduce \textit{Lean4PHYS}, a comprehensive Lean4-based framework for formal physics reasoning. It contains \lib and \bench, where \lib aims to solve the problem that the current Lean4 physics reasoning field lacks foundational support. The \bench, on the other hand, serves as an evaluation method of LLMs' formal physics reasoning capabilities.
    \item Our framework innovates by bridging natural-language physics problems with formal Lean4 representations, enabling LLMs to learn domain-specific laws and reasoning patterns beyond standard math-oriented theorem provers.
    \item We perform extensive experiments by applying leading models to \bench. The experiments suggest that all current expert math provers and general models, regardless of their size, achieve suboptimal performance. The best expert prover achieves \textbf{15.00\%} and the large general model achieves \textbf{39.50\%}. Furthermore, we demonstrate that after integrating \lib, the model exhibits a consistent performance improvement of \textbf{11.75\%}. This suggests that \bench successfully models the limitations of formal physics and the effectiveness of \lib. 
\end{enumerate}

Moreover, to the best of our knowledge, \method is the first work that attempts to extend the LLM-based formal reasoning from math to a more general domain, which offers a new direction that seeks to formalize progressively more subjects. We will open-source \lib and \bench in \url{https://github.com/ShirleyLIYuxin/Lean4PHYS}.
\section{The Lean4PHYS Framework}\label{sec:meth}

In this section, we introduce the design and implementation of the \method framework in detail. The core idea of our framework is to provide a foundation and then evaluate the LLMs' formal physics reasoning capabilities. 
We describe the basic structure of \lib in Section~\ref{meth:lib} and then present the details of how we construct \bench in Section~\ref{meth:bench}.

\subsection{PhysLib}\label{meth:lib}

In this paper, we introduce \lib, a community-driven repository designed to support rigorous and machine-verifiable Lean4 formal physics reasoning. We build the library bottom-up at both the conceptual and technical levels. The current version of \lib contains the foundations of unit systems in physics (Section~\ref{lib:unit}), practical theorems for proving (Section~\ref{lib:topic}), and a guideline for community development and extension (Section~\ref{lib:community}). 

\subsubsection{Foundation of Physics: Unit System}\label{lib:unit}

The central challenge of formalizing physics from scratch is to abstract the kernel that supports all reasoning in the field. Unlike mathematics, where the basic building blocks are defined by pure definition, physics and other natural sciences are primarily supported by empirical rules derived from experiments and lack explicit definitions. For physics, we lay the foundation of reasoning by establishing the unit system.

From an implementation perspective, we build the unit system by extending Mathlib4~\citep{mathlib} and the Lean4 \textit{UnitSystem} kernel~\citep{teorth_analysis}. The unit system contains seven basic units, including \texttt{time}, \texttt{length}, \texttt{mass}, \texttt{electric current}, \texttt{temperature}, \texttt{amount of substance}, and \texttt{unit of luminous intensity}. We define basic \texttt{Normed Space}, \texttt{algebraic computation}, and \texttt{derivative} for the physics unit system in Lean4, which are vital for both problem construction and proof. Additionally, we prove the \texttt{interchangeability} between physics quantities and mathematical quantities. Besides, we also define the \texttt{physics dimension cast} and prove that such a cast does not affect the mathematical propriety of such a unit.

\subsubsection{Topic-based theorem development}\label{lib:topic}

Building on the cross-topic kernel foundation system of units, we introduce topic-based theorem systems tailored to different needs in specific problem types. Specifically, the current version of \lib splits problems into six major topics, namely: \texttt{mechanics}, \texttt{waves \& acoustics}, \texttt{thermodynamics}, \texttt{electromagnetism}, \texttt{optics}, and \texttt{modern physics}. The topic split is inspired by~\citet{young2019university}. 
Our implementation principle for this section of \lib is to first create different namespaces and independent Lean files for each topic. Subsequently, we add topic-specific unit types and constants in the topic namespace. Then, we implement basic physics rules summarized from experiments as definitions. Finally, we implement theorems with their proofs, which are relevant to the topic, as the final layer. We implement the mechanics field in detail as an example and set a basic foundation for other topics. We present the design process to the community and launch this project for collaborative development of the field.

\subsubsection{Community-driven and Extensibility}\label{lib:community}

As mentioned above, \lib is designed to be a community-driven, collaborative work like Mathlib~\citep{mathlib}, and we make our best effort to ensure that other researchers can easily read and extend the system while maintaining consistency. In general, we organize \lib in a three-layer hierarchy: (1) Foundation unit system, which should be consistent with only the necessary changes. (2) Topic-specific unit system, which should be added when formalizing theorem statements if current units are unable to support the construction. (3) Topic-specific theorems, which include most of the practical theorems to support proof implementation and should be regularly updated.

The current version contains a relatively comprehensive implementation of the repository in mechanics, serving as an example for the community. \lib inherits components from~\citet{teorth_analysis}, which are distributed under the Apache-2.0 license, and the same license applies to newly added content. We will actively encourage contributions of new statements and proofs, and will continuously maintain the repository in the future. 

Beyond the current repository we are implementing, such a layered design applies to the formalization of other domains, such as other natural and social sciences, and can be alternatively extended to general proving systems based on the same logic.

\subsection{LeanPhysBench}\label{meth:bench}

Given the current trend of using LLMs for formal reasoning, it is crucial to evaluate their ability to perform formal physics reasoning. However, to the best of our knowledge, there is no benchmark dataset for evaluating LLMs' Lean4 physics reasoning capabilities. Thus, we propose (to the best of our knowledge) the first benchmark for Lean4 physics reasoning. In this section, we detail the process of creating this benchmark. We firstly present the data collection process in Section~\ref{bench:data}, then detail how we created the benchmark in Section~\ref{bench:formalization}, and report the benchmark statistics in Section~\ref{bench:stat}.

\subsubsection{Data Collection \& Preprocessing}\label{bench:data}

Following the bottom-up principle, we build the \bench from the basics to the advanced. The \bench primarily consists of two levels of data: the high school Olympiad competition and the college textbook. The Olympiad data are collected from competition-related exercise books, which cover introductory college-level knowledge with more tedious tricks. It focuses on testing the model's capability to perform multi-step reasoning within a specific field of knowledge. 
On the other hand, the college textbook problems are selected to cover a wider range of concepts with relatively easier reasoning, which focus on testing the LLM's capability to reason across multiple physics models. For questions accompanied by figures, we manually convert this information into Natural Language (NL) and add it as the context to the problems. Detailed data sources are provided in Appendix~\ref{appendix:data}.

Following \lib's design, we further divided the topic of the \bench into mechanics, waves \& acoustics, thermodynamics, electromagnetism, optics, and modern physics. After data collection, we have the base NL statement for formulating \bench.

\subsubsection{Formalization pipeline}\label{bench:formalization}

After collecting and preprocessing the NL problems, we apply a strict pipeline to transform them into verifiable Lean4 theorems. The overview of the formalization pipeline is shown in Figure~\ref{fig:Physbench_Pipeline}. Formally, if we denote a physics problem we want to formalize by $P$, the original problem we have is $P_{original}$, the target is to obtain the Lean4 version of the problem $P_{Lean}$. The formalization process is as follows:

\paragraph{NL Format Alignment} According to previous work in formalizing mathematical statements~\citep{wang2025let, zheng2021minif2f}, there is a significant representation gap between Lean4 statements and their corresponding NL problems in the original datasets. Specifically, in NL problems, it is typical for them to be in Question-Answering (QA) format, where the goal is to find a specific numerical or formulaic answer. However, in Lean4, the problem type is a closed-end proof rather than a specific answer, which creates a gap in the statement. Inspired by~\cite{wang2025let}, we perform a format alignment to transform the QA style physics problem into a proof statement. 

We first transform the NL problem into a proof format. Specifically, we transform the question part of ``Finding the answer to ...'' into ``Prove that the answer is ...'' following~\cite{zheng2021minif2f}. Subsequently, to better model the physical process and relations, we require the LLM to write a step-by-step solution to the QA problem. Based on the solution, we extract the physical laws used in the problems and all the initial conditions to compose statements. Finally, we define the proving target for the problem. We split the target of proof into three categories — numerical value, physical expression, or logical formula describing a physical state — and use these targets to assist the process of Lean4 code writing. After this step, we align the NL question-answering problem with a provable question using extracted physics laws, initial conditions, and a proving target.

\begin{figure}[t]
\centering
\footnotesize
% Two plots side by side
\begin{minipage}[t]{0.49\textwidth}
\centering
\begin{adjustbox}{max width=\linewidth}
\begin{tcolorbox}[title=College Level, colback=white!95!gray, colframe=black!65!gray, boxrule=0.5pt, arc=1mm, top=0mm, bottom=0mm, left=2mm, right=2mm]
\begin{lstlisting}[basicstyle=\tiny]
theorem University_Mechanics_3
  (x_0 x_1 : Length)
  (t_0 t_1 dt t : Time)
  (v_0 v_1 : Speed)
  (a: Acceleration)
  (xf xf1: Time → Length)(vf : Time → Speed)
  (ht0 : t_0 = 0 • second)
  (ht1 : t_1 = 4 • second)
  (ht : dt = t_1 - t_0)
  (ha : a = (v_1 -v_0)/dt)
  (hv : ∀ t, vf t = v_0 + a * t/1)
  (hv1: v_1 = vf dt)
  (hx: ∀ t, xf t = (a * t**2)/2 + v_0 * t)
  (hxx: ∀ t, xf1 t = (3 • meter / second**2)* t**2 - 
    2 • meter / second * t)
  (hxxx: xf = xf1):
  (a = 6 • meter / second**2 ∧ 
   v_0 = -2 • meter / second)
:= by sorry
\end{lstlisting}
\end{tcolorbox}
\end{adjustbox}
\end{minipage}
\hfill
\begin{minipage}[t]{0.49\textwidth}
\centering
\begin{adjustbox}{max width=\linewidth}
\begin{tcolorbox}[title=Competition Level, colback=white!95!gray, colframe=black!65!gray, boxrule=0.5pt, arc=1mm, top=0mm, bottom=0mm, left=2mm, right=2mm]
\begin{lstlisting}[basicstyle=\tiny]
theorem competition_mechanics_Ch2_Q32
  (m : Mass) (R : Length) (θ : ℝ) (v : Speed) (μ : ℝ) (N f : Force)
  (h_pos : 0 < m.val ∧ 0 < μ ∧ 0 < g.val)
  (h_sin_cos : Real.sin θ ≠ 0 ∧ Real.cos θ ≠ 0)
  (r_def : r = Real.sin θ • R)
  (h_horiz : Real.sin θ • N - Real.cos θ • f = 
    m * v**2 / r)
  (h_vert  : Real.cos θ • N + Real.sin θ • f = m * g)
  (f_def : f = m * (Real.sin θ • g - 
    Real.cos θ • v**2 / r / 1))
  (N_def : N = m * (Real.cos θ • g + 
    Real.sin θ • v**2 / r / 1))
  (fric_bound : ‖f.val‖ ≤ μ * ‖N.val‖) :
  ∀ (ε : ℤ), (ε = 1 ∨ ε = -1) →( 
    f = (ε : ℝ) • μ • N → v**2 = 
    ((Real.sin θ - (ε : ℝ) * μ * Real.cos θ) • g * 
    (Real.sin θ • R)/ 
    (Real.cos θ + (ε : ℝ) * μ * Real.sin θ))) 
:= by sorry
\end{lstlisting}
\end{tcolorbox}
\end{adjustbox}
\end{minipage}
\caption{
Two examples from \bench demonstrating different difficulty levels.
% Two examples from \textit{LeanPhysBench} demonstrating different difficulty levels: (top) and Olympiad competition level (bottom). 
% Textbook-level questions focus on simple operations and estimations of physical quantities, while competition-level physics emphasizes derivations of physical formulas and requires stronger mathematical tools.
}
\label{fig:leanphysbench_levels}
\vspace{-0.2in}
\end{figure}

\paragraph{Lean4 code writing \& Verification} 
After we obtain aligned NL problems, the Lean code writing is split into the formalization of conditions and goals. 
If we find any laws missing from \lib that are necessary for us to write the problem, we implement such laws in \lib and represent them correspondingly in our problem.
In formalizing goals, we write the corresponding Lean4 expression of the proving targets as goals for the Lean4 proof. After we obtain the Lean4 code, we submit it to the verifier to check whether it compiles successfully. Additionally, we ask experts in physics and Lean to check the completeness of the theorem. Each statement requires one expert to formalize and at least two experts to verify. 

The above pipeline ensures that the problems in \bench accurately capture the underlying physics semantics from NL descriptions. We present two examples of Lean4 statements we formalized in Figure~\ref{fig:leanphysbench_levels}. In the examples, we can clearly see that the college-level problems focus on relatively simple operations in a broader range of topics. Competition-level statements emphasize derivations of formulas using more advanced math tools in a more concentrated field of physics.

Additionally, during our formalization, based on the number of newly crafted theorems used in the statement, we further divide the competition level problem into easy and hard sub-levels. The easy-level problems use less content from \lib and more from Mathlib4, and are more focused on deriving mathematical formulas with looser conditions. On the other hand, hard-level problems require many theorems and definitions in \lib to answer with tighter conditions. This split aims to evaluate the model's ability to use the OOD theorems provided in the context.

\setlength\intextsep{0pt} 
\setlength\columnsep{10pt} 
\begin{wrapfigure}{r}{0.5\textwidth} 
    \vspace{-0.2in}
    \centering
    \includegraphics[width=0.5\textwidth]{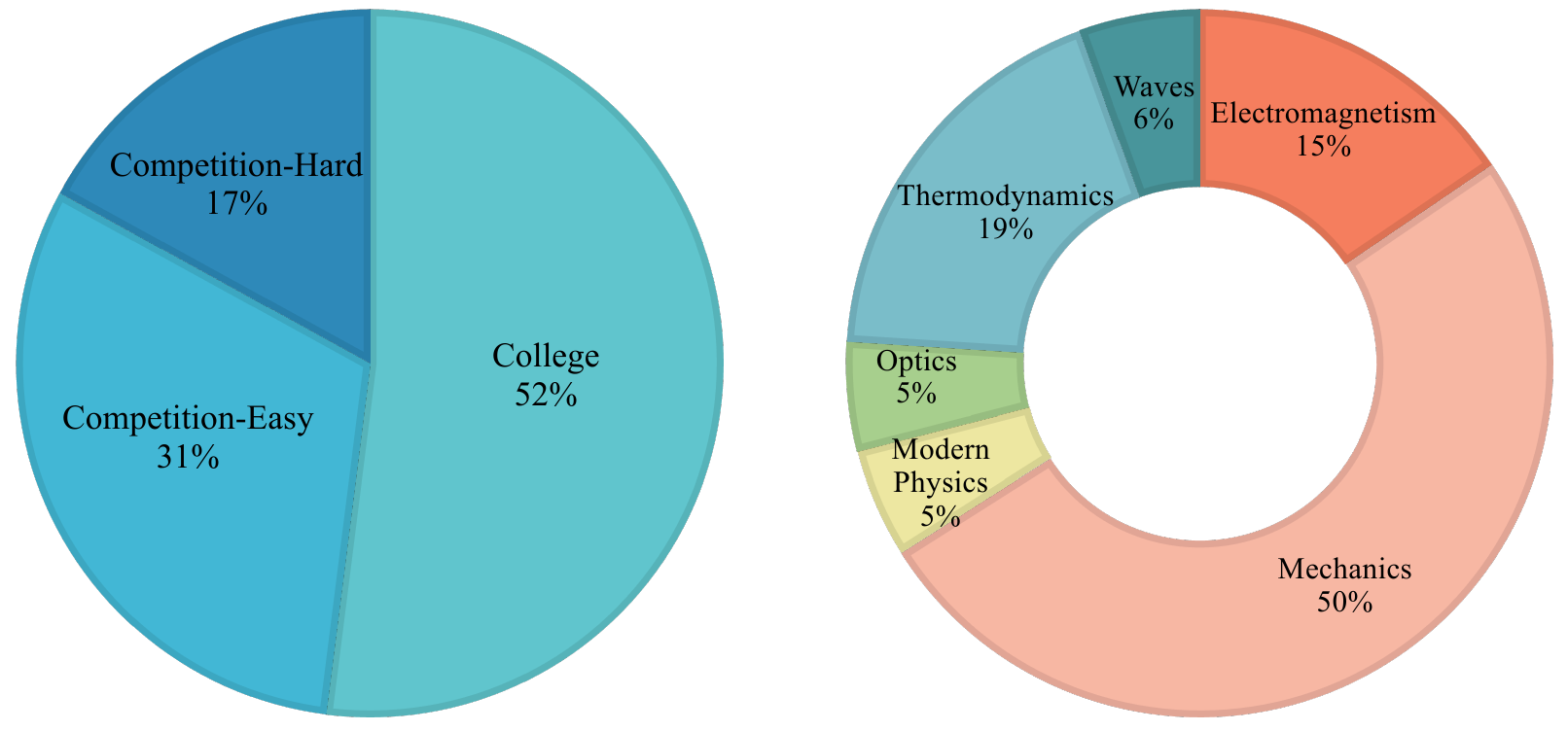}
    \caption{Statistics of \bench: The distribution of 200 Lean4 physics statements across difficulty levels (on the left) and topics (on the right). 
    }
    \label{fig:stat}
    \vspace{-0.2in}
\end{wrapfigure}

\subsubsection{Benchmark Statistics}\label{bench:stat}

Detailed statistics for the \bench are presented in Figure~\ref{fig:stat}. In total, \bench contains 200 physics statements formalized in Lean4. Among them, 104 statements are at the college-level and 96 are at the competition level. The competition level is further divided into 62 easy problems and 34 hard problems. 

\section{Experiment}\label{sec:exp}

We conduct comprehensive experiments on \method to demonstrate the significance of our proposed \bench and the effectiveness of \lib. In particular, we present the performance of major expert Lean provers and general models on \bench and prove the effectiveness of \lib in Section~\ref{exp:results}. We study the problem format of \bench in Section~\ref{exp:format}, and in Section~\ref{exp:case} we perform case studies to address critical concerns raised by main results.

\begin{table*}[t]
    \caption{
    Pass@16 results of \bench on 8 LLMs in with (\ding{51}) and without (\ding{55}) \lib mode across different difficulty levels, including College, Competition-Easy (Comp-Easy), and Comp-Hard. The best result is \textbf{bolded} and the second-best result is \underline{underlined}.
    }
    \resizebox{\textwidth}{!}{
        \begin{tabular}{cccccc}
        \toprule
            \textbf{Method} 
            & \textbf{PhysLib}  & \textbf{College}      & \textbf{Comp-Easy}    & \textbf{Comp-Hard}   & \textbf{Overall} \\
            \midrule
            \multicolumn{6}{l}{\textit{Open-source models}} \\
            \midrule
            \multirow{2}{*}{DeepSeek-R1-8B~\citep{guo2025deepseek}} 
             & \ding{51}        & 6.73\%                & 9.68\%                & 0.00\%                & 6.50\% \\
             & \ding{55}        & 0.00\%                & 4.84\%                & 2.94\%                & 2.00\% \\
            \midrule
            \multirow{2}{*}{Qwen3-8B~\citep{qwen3}} 
             & \ding{51}        & 7.69\%                & 8.06\%                & 0.00\%                & 2.00\% \\
             & \ding{55}        & 1.92\%                & 3.23\%                & 0.00\%                & 6.50\% \\
            \midrule
            \multirow{2}{*}{Kimina-Prover-8B~\citep{wang2025kimina}} 
             & \ding{51}        & 8.65\%                & 20.97\%               & \textbf{8.82\%}       & 12.50\% \\
             & \ding{55}        & 5.77\%                & 16.13\%               & \underline{5.88\%}    & 9.00\% \\
            \midrule
            \multirow{2}{*}{Goedel-Prover-V2-8B~\citep{lin2025goedel}} 
             & \ding{51}        & 8.65\%                & 24.19\%               & 2.94\%                & 13.00\% \\
             & \ding{55}        & 6.73\%                & 19.35\%               & \underline{5.88\%}    & 10.00\% \\
            \midrule
            \multirow{2}{*}{DeepSeek-Prover-V2-7B~\citep{ren2025deepseek}} 
             & \ding{51}        & 8.65\%                & 29.03\%               & \underline{5.88\%}    & 14.50\% \\
             & \ding{55}        & 6.73\%                & 22.58\%               & \underline{5.88\%}    & 11.50\% \\
            \midrule
            \multicolumn{6}{l}{\textit{Closed-source models}} \\
            \midrule
            \multirow{2}{*}{GPT-4o~\citep{openai2024gpt}} 
             & \ding{51}        & 6.73\%                & 30.65\%               & 0.00\%                & 13.00\% \\
             & \ding{55}        & 2.88\%                & 1.61\%                & 0.00\%                & 2.00\% \\
            \midrule
            \multirow{2}{*}{Claude-Sonnet-4~\citep{anthropic_claude_sonnet_4}} 
             & \ding{51}        & \underline{28.85\%}   & \underline{62.90\%}   & 0.00\%                & \underline{34.50\%} \\
             & \ding{55}        & 0.96\%                & 4.84\%                & 0.00\%                & 2.00\% \\
            \midrule
            \multirow{2}{*}{Gemini-2.5-pro~\citep{comanici2025gemini}} 
             & \ding{51}        & \textbf{31.73\%}      & \textbf{74.19\%}      & 0.00\%                & \textbf{39.50\%} \\
             & \ding{55}        & 6.73\%                & 12.90\%               & 0.00\%                & 7.50\% \\
        \bottomrule
        \end{tabular}
    }
    \centering
    \small
    \label{tab:main}
\vspace{-0.15in}
\end{table*}

\subsection{Experiment Setup}\label{exp:setup}

We evaluate the LLMs' physics-reasoning capabilities in Lean4 by applying them to write proofs for the \bench. Specifically, the task for the LLM is to write Lean4 proofs with NL statements and Lean4 statements of the physics problems in the prompt. Following the standard in~\cite{xin2024deepseek, wang2024theoremllama}, we manually configure all the imports and namespaces. Furthermore, unless otherwise specified, we allow the LLM to perform Long Chain-of-Thought thinking to perform deeper reasoning. 

To better demonstrate current LLMs' formal physical reasoning capability, we select the most representative closed-source and open-source models to evaluate. Namely, for closed-source general LLMs, we select GPT-4o~\citep{openai2024gpt}, Claude-Sonnet-4~\citep{anthropic_claude_sonnet_4}, and Gemini-2.5-Pro~\citep{comanici2025gemini} as baselines. For open-source models, we present results from general-purpose models, including DeepSeek-R1-0528~\citep{guo2025deepseek} and Qwen3-8B~\citep{qwen3}, as well as expert Lean4 provers such as Goedel-Prover-V2-8B~\citep{lin2025goedel2}, Kimina-Prover~\citep{wang2025kimina}, and DeepSeek-Prover-V2~\citep{ren2025deepseek}. 

Furthermore, to demonstrate the effectiveness of \lib, we test the LLM's capability under modes that have or do not have \lib in the generated context. In summary, we test the \bench on eight major LLMs under two modes. The implementation details of our experiments can be found in Appendix~\ref{appendix:exp_setup}.

\subsection{Main Result}\label{exp:results}

The main experiment results are demonstrated in Table~\ref{tab:main}. From the table, we can see that all models suffer from suboptimal overall performance, with the leading open-source models achieving at most 14.50\%, while large general models achieve a high accuracy of 40.50\% across the entire dataset, which is also relatively lower than their performance in NL math tasks of a similar difficulty level. Furthermore, in the context of \lib, Gemini achieves higher accuracy across the entire dataset, while the performance enhancement for smaller models is less significant. It indicates the effectiveness of our \lib in assessing the LLMs' formal physics reasoning.

Upon closer examination, we can observe that Lean experts, which significantly outperform closed-source general LLMs in the math domain, lack the strong formal physics capabilities of these models. This reveals that the Lean expert provers' capabilities are limited to the math domain and hard to transfer to physics. Especially when these domains apply a new definition (such as unit system in \method). Moreover, we found that the performance difference between expert provers is relatively marginal, indicating that significant improvement in mathematics does not necessarily translate into a large improvement in physics reasoning. 

When analyzing results across different levels of difficulty, we find that the models generally perform well on easy competition problems, which are the most closely aligned with mathematical representations. However, expert models still do not outperform larger, closed-source models, indicating their limited ability to transfer mathematical knowledge into physics theorem proofs. This finding also holds for college-level problems, where the expert model's performance is generally lower. However, on the competition-hard level of problems, the expert provers perform better than the closed-source models, which achieve zero performance. This is because the expert models have a stronger capability to perform extremely long and complex deductions, whereas large models do not.

Furthermore, adding \lib to the context will significantly improve the LLMs' performance in almost all cases. 
We conclude that this improvement stems from the better in-context understanding of the unit system and the more accurate selection of relevant theorems. In contrast, without the \lib, the model can only perform basic simplification tactics like: \texttt{constructor}, \texttt{rw}, \texttt{abel}, \texttt{exact}, \texttt{aesop}. By adding the \lib, the model can learn to perform more advanced tactics, such as \texttt{simp} and \texttt{norm\_num}.

\subsection{Problem Format Study}\label{exp:format}

This section presents a more detailed study of the problem format. Due to the space limit, we place the example in Figure~\ref{fig:PROBLEM_FORMAT_STUDY} in the Appendix and only provide the analysis result here.

From the example of problem format, we observe that college-level statements primarily involve numeric computations with a relatively wide range of physical quantities with units. These problems rarely require multi-step formula derivations, but heavily depend on the unit system in \lib to ensure dimensional consistency. Thus, the models with weaker in-context learning perform relatively poorly on this level of problems. This is because they cannot infer the new out-of-distribution syntax or unit-handling rules from context.

The easy-level competition questions are closer to traditional math problems in Lean, such as MiniF2F~\citep{zheng2021minif2f}. They involve relatively simple formula derivations, typically within two steps and often only tactics from Mathlib. Therefore, the models that are more familiar with the Lean mathematics and good at using tactics perform better than closed-source large models on these kinds of problems even without \lib. 

On the other hand, the hard level of competition problems demands complex symbolic reasoning, handling quantifiers, and difficult functional reasoning. For instance, this includes proving the existence of a number, holding an inequality, or deriving a functional relation. These problems combine unit casts with symbolic manipulation, further increasing the difficulty. Solving this level of problem requires careful decomposition of the problem, proving new lemmas, and multiple proof strategies with creativity. Moreover, many problems of this level require calculus concepts such as limits and continuity. All of the above factors combined cause the low pass rate at this level.

\subsection{Case Study}\label{exp:case}

We present the case analysis in this section to provide a more detailed examination of the key findings from our main experiment. Due to the space limit, for a detailed example, please refer to Appendix~\ref{appendix:case_study}.

\paragraph{Behavior of the same theorem with and without \lib} Figure~\ref{fig:case_differentproofs} demonstrates a theorem at the college level, which Gemini can do both with and without \lib in context. From the comparison, we find that without the library, the proof is based solely on Mathlib. When the \lib is in the context, the proof tends to use the operations in the library and include explanatory comments. It indicates \lib can assist LLM's reasoning by providing a wider toolbox.

\paragraph{Transfer tricks in mathematics} Figure~\ref{fig:goedel-only} demonstrate a problem only solved by Goedel-Prover-V2-8B in our entire cycle of experiment. We conclude that this is a good transfer of following the NL intermediate steps and performing multiple trials learned in the math Lean training. From the NL statement demonstrated in Figure~\ref{fig:goedel-only-1}, we observe that most steps are presented in the statement. Following these NL hints and many trials using \texttt{try} tactics, the Goedel-Provers can successfully solve the problem. It indicates that, although limited, some of the expert prover's capability of Lean math reasoning can be applied to formal physics reasoning.

\paragraph{Why the general model performs better} To answer this question, we further analyze different topics solved by LLMs and find that general LLMs are typically better in thermo-dynamics. The Gemini successfully proved 22 theorems in the field, but DeepSeek-Prover-V2 only finished six theorems. We present one theorem that is proved by both Gemini-2.5 and DeepSeek-Prover-V2 to study the different behavior of their proof in Figure~\ref{fig:case_proofstyle}. We observe that the expert prover's proof is more complex and tedious, whereas Gemini's is cleaner and more straightforward. Such a difference in proving consistently shows between the expert prover and general models. It suggests that in fields underrepresented in training, the prolonged deliberation of expert models may lead to overthinking and result in suboptimal results, a problem that may be addressed by adaptive reasoning techniques~\citep{huang2025adactrl} that dynamically adjust the LLM reasoning strategy.
\section{Related Work}\label{sec:relat}

\subsection{Formal Reasoning}\label{relat:formal}

Formal Language (FL) reasoning involves expressing mathematical statements in a computer-verifiable manner. This approach mitigates ambiguity and provides a solid foundation for the reasoning process. Researchers have developed many FLs in the last decades, such as Isabelle~\citep{paulson1994isabelle}, CoQ~\citep{coq1996coq}, Metamath~\citep{megill2019metamath}, and Lean~\citep{de2015lean}. Among these, Lean4~\citep{moura2021lean} receives significant attention from the field due to its extensive foundation library of Mathlib~\citep{mathlib}. A series of datasets and benchmarks has been developed to advance LLM Lean4 reasoning. For example, MiniF2F~\citep{zheng2021minif2f} formalized competition-level math problems across multiple proof languages, ProofNet~\citep{azerbayev2023proofnet} increases the level of difficulty to college-level, and PutnamBench~\citep{gulati2024putnam} serves as college-level competition problems, with MATPBench~\citep{he2025matpbenchmllmgoodautomated} taking a step further by extending formal proof into multi-modal reasoning. Meanwhile, large-scale datasets have also been proposed to support LLM training, pushing the training data from 100k level~\citep{wang2024theoremllama} to millions level~\citep{lin2025goedel, dong2025stp}. At the same time, a series of Lean expert models are developed to perform formal math reasoning, representing works like DeepSeek-Prover Family~\citep{xin2024deepseek, ren2025deepseek}, Goedel-Prover Family~\citep{lin2025goedel, lin2025goedel2}, LoT-Solver~\citep{wang2025ma}, Kimina-Prover~\citep{wang2025kimina}, BFS-Prover~\citep{xin2025bfs}, and TheoremLlama~\citep{wang2024theoremllama}.

\subsection{Lean in Subjects Beyond Mathematics}\label{relat:byd}
The application of Lean is gradually expanding from mathematics to a wider range of scientific fields, including chemical physics~\citep{bobbin2024formalizing}, molecular simulation~\citep{ugwuanyi2025benchmarking}, and electrical engineering~\citep{blacksph3re2025distribution_factors}. Some projects, such as ~\citet{tooby2025heplean}, attempt to formalize mechanics and high-energy physics using tools such as tensors and index notation. However, most of these studies are still limited to a small-scale, non-modular level for specific theorems, and lack a standardized evaluation system. Although the Lean expert provers perform well on mathematical benchmarks, their ability to transfer their reasoning skills to other fields, such as physics, remains to be examined.

\subsection{Physics Datasets}\label{relat:phys}
In the field of physical reasoning research, the construction of data sets has long been a core issue. Physical problems have been extensively studied in the fields of natural language (NL) and machine learning, giving rise to several benchmarks that focus on physical understanding and reasoning. For instance, PhysBench~\citep{chow2025physbenchbenchmarkingenhancingvisionlanguage} and PHYBench~\citep{qiu2025phybenchholisticevaluationphysical} evaluate the physical reasoning ability of models; at the course level, UG-Physics~\citep{xu2025ugphysics} and the PHYSICS dataset~\citep{zheng2025physicsdataset} provide thousands of textbook-style questions; at the competition level, OlympiadBench~\citep{he2024olympiadbench} includes 8,476 bilingual physics competition questions, HiPhO~\citep{yu2025hipho} covers the recent international physics olympiad questions, and PhysReason~\citep{zhang2025physreason} emphasizes the combination of multi-step reasoning and image understanding. Furthermore, CAMEL-Physics~\citep{li2023camel} expanded the scale to tens of thousands of questions through automatic generation technology, providing a broader resource for model evaluation and training. Overall, these multi-level physical datasets laid the foundation for testing the physical reasoning ability of large language models under natural language conditions.

\section{Conclusion}\label{sec:conclusion}

This paper presents \method, a comprehensive framework to support Lean4 physics reasoning. 
The framework includes \lib, an extensible, community-driven foundation library that sets the cornerstone for units, fields, and theorems for formal physics reasoning. 
To evaluate LLMs’ performance on formal physics reasoning, we propose \bench, which is, to the best of our knowledge, the first benchmark for Lean4-based physics reasoning. 
Based on the \method, we conduct extensive experiments to provide an overview of LLMs' performance on such tasks and the effectiveness of our \lib. This reveals that current models exhibit suboptimal performance in Lean4 physics reasoning, whereas open-source expert provers do not outperform closed-source general models in formal physics reasoning. This indicates limited transfer capability from mathematical reasoning, despite the fact that they are all Lean4-based. Furthermore, the experiment shows that with \lib in the context, LLMs' performance on \bench increases by \textbf{11.75\%} on average. Our work establishes a general principle for extending the formalization of physics and other natural sciences beyond mathematics into a verifiable system.

\section*{Acknowledgments}
We would like to express our gratitude to Yifei Xia for her valuable assistance in figure plotting and helpful comments during the paper writing process. We thank Wenyuan Wang for sharing his precious experience in the engineering aspects of formal verification. We also thank Peng Chen for generously providing original collections of high school physics problems, which served as valuable references during the design phase of \method.  
\bibliography{iclr2026_conference}

\begin{thebibliography}{53}
\providecommand{\natexlab}[1]{#1}
\providecommand{\url}[1]{\texttt{#1}}
\expandafter\ifx\csname urlstyle\endcsname\relax
  \providecommand{\doi}[1]{doi: #1}\else
  \providecommand{\doi}{doi: \begingroup \urlstyle{rm}\Url}\fi

\bibitem[Anthropic(2025)]{anthropic_claude_sonnet_4}
Anthropic.
\newblock Claude sonnet 4, 2025.
\newblock URL \url{https://www.anthropic.com/claude/sonnet}.
\newblock Accessed: 2025-09-23.

\bibitem[Azerbayev et~al.(2023)Azerbayev, Piotrowski, Schoelkopf, Ayers, Radev, and Avigad]{azerbayev2023proofnet}
Zhangir Azerbayev, Bartosz Piotrowski, Hailey Schoelkopf, Edward~W Ayers, Dragomir Radev, and Jeremy Avigad.
\newblock Proofnet: Autoformalizing and formally proving undergraduate-level mathematics.
\newblock \emph{arXiv preprint arXiv:2302.12433}, 2023.

\bibitem[blacksph3re(2025)]{blacksph3re2025distribution_factors}
blacksph3re.
\newblock distribution\_factors: An attempt to formalize dc loadflow equations and distribution factors in lean, 2025.
\newblock URL \url{https://github.com/blacksph3re/distribution_factors}.
\newblock Accessed: 2025-09-25.

\bibitem[Bobbin et~al.(2024)Bobbin, Sharlin, Feyzishendi, Dang, Wraback, and Josephson]{bobbin2024formalizing}
Maxwell~P Bobbin, Samiha Sharlin, Parivash Feyzishendi, An~Hong Dang, Catherine~M Wraback, and Tyler~R Josephson.
\newblock Formalizing chemical physics using the lean theorem prover.
\newblock \emph{Digital Discovery}, 3\penalty0 (2):\penalty0 264--280, 2024.

\bibitem[Chow et~al.(2025)Chow, Mao, Li, Seita, Guizilini, and Wang]{chow2025physbenchbenchmarkingenhancingvisionlanguage}
Wei Chow, Jiageng Mao, Boyi Li, Daniel Seita, Vitor Guizilini, and Yue Wang.
\newblock Physbench: Benchmarking and enhancing vision-language models for physical world understanding, 2025.
\newblock URL \url{https://arxiv.org/abs/2501.16411}.

\bibitem[Cobbe et~al.(2021)Cobbe, Kosaraju, Bavarian, Chen, Jun, Kaiser, Plappert, Tworek, Hilton, Nakano, Hesse, and Schulman]{cobbe2021gsm8k}
Karl Cobbe, Vineet Kosaraju, Mohammad Bavarian, Mark Chen, Heewoo Jun, Lukasz Kaiser, Matthias Plappert, Jerry Tworek, Jacob Hilton, Reiichiro Nakano, Christopher Hesse, and John Schulman.
\newblock Training verifiers to solve math word problems.
\newblock \emph{arXiv preprint arXiv:2110.14168}, 2021.

\bibitem[Comanici et~al.(2025)Comanici, Bieber, Schaekermann, Pasupat, Sachdeva, Dhillon, Blistein, Ram, Zhang, Rosen, et~al.]{comanici2025gemini}
Gheorghe Comanici, Eric Bieber, Mike Schaekermann, Ice Pasupat, Noveen Sachdeva, Inderjit Dhillon, Marcel Blistein, Ori Ram, Dan Zhang, Evan Rosen, et~al.
\newblock Gemini 2.5: Pushing the frontier with advanced reasoning, multimodality, long context, and next generation agentic capabilities.
\newblock \emph{arXiv preprint arXiv:2507.06261}, 2025.

\bibitem[Coq(1996)]{coq1996coq}
Projet Coq.
\newblock The coq proof assistant-reference manual.
\newblock \emph{INRIA Rocquencourt and ENS Lyon, version}, 5, 1996.

\bibitem[De~Moura et~al.(2015)De~Moura, Kong, Avigad, Van~Doorn, and von Raumer]{de2015lean}
Leonardo De~Moura, Soonho Kong, Jeremy Avigad, Floris Van~Doorn, and Jakob von Raumer.
\newblock The lean theorem prover (system description).
\newblock In \emph{Automated Deduction-CADE-25: 25th International Conference on Automated Deduction, Berlin, Germany, August 1-7, 2015, Proceedings 25}, pp.\  378--388. Springer, 2015.

\bibitem[Dong \& Ma(2025)Dong and Ma]{dong2025stp}
Kefan Dong and Tengyu Ma.
\newblock Stp: Self-play llm theorem provers with iterative conjecturing and proving, 2025.
\newblock URL \url{https://arxiv.org/abs/2502.00212}.

\bibitem[Edwards et~al.(2025)Edwards, Han, Lee, Nguyen, Jin, Prasad, Szymku{\'c}, Grzybowski, Diao, Han, et~al.]{edwards2025mclm}
Carl Edwards, Chi Han, Gawon Lee, Thao Nguyen, Bowen Jin, Chetan~Kumar Prasad, Sara Szymku{\'c}, Bartosz~A Grzybowski, Ying Diao, Jiawei Han, et~al.
\newblock mclm: A function-infused and synthesis-friendly modular chemical language model.
\newblock \emph{arXiv preprint arXiv:2505.12565}, 2025.

\bibitem[Gulati et~al.(2024)Gulati, Miranda, Chen, Xia, Fronsdal, de~Moraes~Dumont, and Koyejo]{gulati2024putnam}
Aryan Gulati, Brando Miranda, Eric Chen, Emily Xia, Kai Fronsdal, Bruno de~Moraes~Dumont, and Sanmi Koyejo.
\newblock Putnam-axiom: A functional and static benchmark for measuring higher level mathematical reasoning.
\newblock In \emph{The 4th Workshop on Mathematical Reasoning and AI at NeurIPS'24}, 2024.

\bibitem[Guo et~al.(2025)Guo, Yang, Zhang, Song, Zhang, Xu, Zhu, Ma, Wang, Bi, et~al.]{guo2025deepseek}
Daya Guo, Dejian Yang, Haowei Zhang, Junxiao Song, Ruoyu Zhang, Runxin Xu, Qihao Zhu, Shirong Ma, Peiyi Wang, Xiao Bi, et~al.
\newblock Deepseek-r1: Incentivizing reasoning capability in llms via reinforcement learning.
\newblock \emph{arXiv preprint arXiv:2501.12948}, 2025.

\bibitem[Harrison(2009)]{harrison2009hol}
John Harrison.
\newblock Hol light: An overview.
\newblock In \emph{International Conference on Theorem Proving in Higher Order Logics}, pp.\  60--66. Springer, 2009.

\bibitem[He et~al.(2024)He, Luo, Bai, Hu, Thai, Shen, Hu, Han, Huang, Zhang, et~al.]{he2024olympiadbench}
Chaoqun He, Renjie Luo, Yuzhuo Bai, Shengding Hu, Zhen~Leng Thai, Junhao Shen, Jinyi Hu, Xu~Han, Yujie Huang, Yuxiang Zhang, et~al.
\newblock Olympiadbench: A challenging benchmark for promoting agi with olympiad-level bilingual multimodal scientific problems.
\newblock \emph{arXiv preprint arXiv:2402.14008}, 2024.

\bibitem[He et~al.(2025{\natexlab{a}})He, Qian, Chen, He, Fung, and Ji]{he2025verir1precisefaithfulclaim}
Qi~He, Cheng Qian, Xiusi Chen, Bingxiang He, Yi~R. Fung, and Heng Ji.
\newblock Veri-r1: Toward precise and faithful claim verification via online reinforcement learning, 2025{\natexlab{a}}.
\newblock URL \url{https://arxiv.org/abs/2510.01932}.

\bibitem[He et~al.(2025{\natexlab{b}})He, Lyu, Chen, Guo, and Fung]{he2025matpbenchmllmgoodautomated}
Zhitao He, Zongwei Lyu, Dazhong Chen, Dadi Guo, and Yi~R Fung.
\newblock Matp-bench: Can mllm be a good automated theorem prover for multimodal problems?
\newblock \emph{arXiv preprint arXiv:2506.06034}, 2025{\natexlab{b}}.

\bibitem[Hendrycks et~al.(2021)Hendrycks, Burns, Kadavath, Arora, Basart, Tang, Song, and Steinhardt]{hendrycks2021measuring}
Dan Hendrycks, Collin Burns, Saurav Kadavath, Akul Arora, Steven Basart, Eric Tang, Dawn Song, and Jacob Steinhardt.
\newblock Measuring mathematical problem solving with the math dataset.
\newblock \emph{arXiv preprint arXiv:2103.03874}, 2021.

\bibitem[Huang et~al.(2025)Huang, Wang, Zhong, Su, Feng, Cao, and Fung]{huang2025adactrl}
Shijue Huang, Hongru Wang, Wanjun Zhong, Zhaochen Su, Jiazhan Feng, Bowen Cao, and Yi~R Fung.
\newblock Adactrl: Towards adaptive and controllable reasoning via difficulty-aware budgeting.
\newblock \emph{arXiv preprint arXiv:2505.18822}, 2025.

\bibitem[Li et~al.(2023)Li, Hammoud, Itani, Khizbullin, and Ghanem]{li2023camel}
Guohao Li, Hasan Abed Al~Kader Hammoud, Hani Itani, Dmitrii Khizbullin, and Bernard Ghanem.
\newblock Camel: Communicative agents for "mind" exploration of large scale language model society, 2023.

\bibitem[Lin et~al.(2025{\natexlab{a}})Lin, Tang, Lyu, Wu, Lin, Yang, Li, Xia, Chen, Arora, and Jin]{lin2025goedel}
Yong Lin, Shange Tang, Bohan Lyu, Jiayun Wu, Hongzhou Lin, Kaiyu Yang, Jia Li, Mengzhou Xia, Danqi Chen, Sanjeev Arora, and Chi Jin.
\newblock Goedel-prover: A frontier model for open-source automated theorem proving, 2025{\natexlab{a}}.
\newblock URL \url{https://arxiv.org/abs/2502.07640}.

\bibitem[Lin et~al.(2025{\natexlab{b}})Lin, Tang, Lyu, Yang, Chung, Zhao, Jiang, Geng, Ge, Sun, et~al.]{lin2025goedel2}
Yong Lin, Shange Tang, Bohan Lyu, Ziran Yang, Jui-Hui Chung, Haoyu Zhao, Lai Jiang, Yihan Geng, Jiawei Ge, Jingruo Sun, et~al.
\newblock Goedel-prover-v2: Scaling formal theorem proving with scaffolded data synthesis and self-correction.
\newblock \emph{arXiv preprint arXiv:2508.03613}, 2025{\natexlab{b}}.

\bibitem[mathlib Community(2020)]{mathlib}
The mathlib Community.
\newblock The lean mathematical library.
\newblock In \emph{Proceedings of the 9th ACM SIGPLAN International Conference on Certified Programs and Proofs}, CPP 2020, pp.\  367–381, New York, NY, USA, 2020. Association for Computing Machinery.
\newblock ISBN 9781450370974.
\newblock \doi{10.1145/3372885.3373824}.
\newblock URL \url{https://doi.org/10.1145/3372885.3373824}.

\bibitem[Megill \& Wheeler(2019)Megill and Wheeler]{megill2019metamath}
Norman Megill and David~A Wheeler.
\newblock \emph{Metamath: a computer language for mathematical proofs}.
\newblock Lulu. com, 2019.

\bibitem[Moura \& Ullrich(2021)Moura and Ullrich]{moura2021lean}
Leonardo~de Moura and Sebastian Ullrich.
\newblock The lean 4 theorem prover and programming language.
\newblock In \emph{Automated Deduction--CADE 28: 28th International Conference on Automated Deduction, Virtual Event, July 12--15, 2021, Proceedings 28}, pp.\  625--635. Springer, 2021.

\bibitem[OpenAI et~al.(2024)OpenAI, Lerer, Goucher, Perelman, Ramesh, Clark, Ostrow, Welihinda, Hayes, Radford, et~al.]{openai2024gpt}
Aaron~Hurst OpenAI, Adam Lerer, Adam~P Goucher, Adam Perelman, Aditya Ramesh, Aidan Clark, AJ~Ostrow, Akila Welihinda, Alan Hayes, Alec Radford, et~al.
\newblock Gpt-4o system card.
\newblock \emph{arXiv preprint arXiv:2410.21276}, 1\penalty0 (2):\penalty0 3, 2024.

\bibitem[Paulson(1994)]{paulson1994isabelle}
Lawrence~C Paulson.
\newblock \emph{Isabelle: A generic theorem prover}.
\newblock Springer, 1994.

\bibitem[Polu et~al.(2022)Polu, Han, Zheng, Baksys, Babuschkin, and Sutskever]{polu2022formal}
Stanislas Polu, Jesse~Michael Han, Kunhao Zheng, Mantas Baksys, Igor Babuschkin, and Ilya Sutskever.
\newblock Formal mathematics statement curriculum learning.
\newblock \emph{arXiv preprint arXiv:2202.01344}, 2022.

\bibitem[Qiu et~al.(2025)Qiu, Guo, Song, Sun, Cai, Wei, Luo, Yin, Zhang, Hu, Wang, Tang, Chang, Liu, Zhou, Zhang, Zhang, Liu, Li, Zhang, Jing, Yin, Ren, Fu, Ji, Wang, Tian, Lv, Man, Li, Tao, Sun, Liang, Mu, Li, Zhang, Zhang, Li, Xia, Lin, Shen, Chen, Xiong, Wang, Wang, Ni, Zhang, Cui, Shao, Cao, xing Luo, Yang, Zhang, and Zhu]{qiu2025phybenchholisticevaluationphysical}
Shi Qiu, Shaoyang Guo, Zhuo-Yang Song, Yunbo Sun, Zeyu Cai, Jiashen Wei, Tianyu Luo, Yixuan Yin, Haoxu Zhang, Yi~Hu, Chenyang Wang, Chencheng Tang, Haoling Chang, Qi~Liu, Ziheng Zhou, Tianyu Zhang, Jingtian Zhang, Zhangyi Liu, Minghao Li, Yuku Zhang, Boxuan Jing, Xianqi Yin, Yutong Ren, Zizhuo Fu, Jiaming Ji, Weike Wang, Xudong Tian, Anqi Lv, Laifu Man, Jianxiang Li, Feiyu Tao, Qihua Sun, Zhou Liang, Yushu Mu, Zhongxuan Li, Jing-Jun Zhang, Shutao Zhang, Xiaotian Li, Xingqi Xia, Jiawei Lin, Zheyu Shen, Jiahang Chen, Qiuhao Xiong, Binran Wang, Fengyuan Wang, Ziyang Ni, Bohan Zhang, Fan Cui, Changkun Shao, Qing-Hong Cao, Ming xing Luo, Yaodong Yang, Muhan Zhang, and Hua~Xing Zhu.
\newblock Phybench: Holistic evaluation of physical perception and reasoning in large language models, 2025.
\newblock URL \url{https://arxiv.org/abs/2504.16074}.

\bibitem[Ren et~al.(2025)Ren, Shao, Song, Xin, Wang, Zhao, Zhang, Fu, Zhu, Yang, et~al.]{ren2025deepseek}
ZZ~Ren, Zhihong Shao, Junxiao Song, Huajian Xin, Haocheng Wang, Wanjia Zhao, Liyue Zhang, Zhe Fu, Qihao Zhu, Dejian Yang, et~al.
\newblock Deepseek-prover-v2: Advancing formal mathematical reasoning via reinforcement learning for subgoal decomposition.
\newblock \emph{arXiv preprint arXiv:2504.21801}, 2025.

\bibitem[Saikh et~al.(2022)Saikh, Ghosal, Mittal, Ekbal, and Bhattacharyya]{saikh2022scienceqa}
Tanik Saikh, Tirthankar Ghosal, Amish Mittal, Asif Ekbal, and Pushpak Bhattacharyya.
\newblock Scienceqa: A novel resource for question answering on scholarly articles.
\newblock \emph{International Journal on Digital Libraries}, 23\penalty0 (3):\penalty0 289--301, 2022.

\bibitem[Shu et~al.(2008)Shu, Hu, and Chen]{shu2008physics}
Yousheng Shu, Wangyu Hu, and Bingqian Chen.
\newblock \emph{Selected Advanced Physics Problems, Volume II}.
\newblock Science Press, Beijing, 2008.
\newblock ISBN 9787030193563.

\bibitem[Su et~al.(2025)Su, Xia, Guo, Liu, Ma, Qu, Liu, Li, Zeng, Yang, Li, Cheng, Ji, He, and Fung]{su2025thinkingimagesmultimodalreasoning}
Zhaochen Su, Peng Xia, Hangyu Guo, Zhenhua Liu, Yan Ma, Xiaoye Qu, Jiaqi Liu, Yanshu Li, Kaide Zeng, Zhengyuan Yang, Linjie Li, Yu~Cheng, Heng Ji, Junxian He, and Yi~R. Fung.
\newblock Thinking with images for multimodal reasoning: Foundations, methods, and future frontiers, 2025.
\newblock URL \url{https://arxiv.org/abs/2506.23918}.

\bibitem[Suzgun et~al.(2022)Suzgun, Scales, Sch{\"a}rli, Gehrmann, Tay, Chung, Chowdhery, Le, Chi, Zhou, et~al.]{suzgun2022challenging}
Mirac Suzgun, Nathan Scales, Nathanael Sch{\"a}rli, Sebastian Gehrmann, Yi~Tay, Hyung~Won Chung, Aakanksha Chowdhery, Quoc~V Le, Ed~H Chi, Denny Zhou, et~al.
\newblock Challenging big-bench tasks and whether chain-of-thought can solve them.
\newblock \emph{arXiv preprint arXiv:2210.09261}, 2022.

\bibitem[Talmor et~al.(2018)Talmor, Herzig, Lourie, and Berant]{talmor2018commonsenseqa}
Alon Talmor, Jonathan Herzig, Nicholas Lourie, and Jonathan Berant.
\newblock Commonsenseqa: A question answering challenge targeting commonsense knowledge.
\newblock \emph{arXiv preprint arXiv:1811.00937}, 2018.

\bibitem[Tao(2025)]{teorth_analysis}
Terence Tao.
\newblock A lean companion to analysis i.
\newblock \url{https://github.com/teorth/analysis}, 2025.
\newblock URL \url{https://github.com/teorth/analysis}.
\newblock GitHub repository, accessed: 2025-08-31.

\bibitem[Team(2025)]{qwen3}
Qwen Team.
\newblock Qwen3, April 2025.
\newblock URL \url{https://qwenlm.github.io/blog/qwen3/}.

\bibitem[Tooby-Smith(2025)]{tooby2025heplean}
Joseph Tooby-Smith.
\newblock Heplean: Digitalising high energy physics.
\newblock \emph{Computer Physics Communications}, 308:\penalty0 109457, 2025.

\bibitem[Ugwuanyi et~al.(2025)Ugwuanyi, Jones, Velkey, and Josephson]{ugwuanyi2025benchmarking}
Ejike~D Ugwuanyi, Colin~T Jones, John Velkey, and Tyler~R Josephson.
\newblock Benchmarking energy calculations using formal proofs.
\newblock \emph{Molecular Physics}, pp.\  e2539421, 2025.

\bibitem[Wang et~al.(2025{\natexlab{a}})Wang, Unsal, Lin, Baksys, Liu, Santos, Sung, Vinyes, Ying, Zhu, et~al.]{wang2025kimina}
Haiming Wang, Mert Unsal, Xiaohan Lin, Mantas Baksys, Junqi Liu, Marco~Dos Santos, Flood Sung, Marina Vinyes, Zhenzhe Ying, Zekai Zhu, et~al.
\newblock Kimina-prover preview: Towards large formal reasoning models with reinforcement learning.
\newblock \emph{arXiv preprint arXiv:2504.11354}, 2025{\natexlab{a}}.

\bibitem[Wang et~al.(2024{\natexlab{a}})Wang, Zhang, Jia, Pan, Diao, Pi, and Zhang]{wang2024theoremllama}
Ruida Wang, Jipeng Zhang, Yizhen Jia, Rui Pan, Shizhe Diao, Renjie Pi, and Tong Zhang.
\newblock Theoremllama: Transforming general-purpose llms into lean4 experts.
\newblock \emph{arXiv preprint arXiv:2407.03203}, 2024{\natexlab{a}}.

\bibitem[Wang et~al.(2025{\natexlab{b}})Wang, Li, Fung, and Zhang]{wang2025let}
Ruida Wang, Yuxin Li, Yi~R Fung, and Tong Zhang.
\newblock Let's reason formally: Natural-formal hybrid reasoning enhances llm's math capability.
\newblock \emph{arXiv preprint arXiv:2505.23703}, 2025{\natexlab{b}}.

\bibitem[Wang et~al.(2025{\natexlab{c}})Wang, Pan, Li, Zhang, Jia, Diao, Pi, Hu, and Zhang]{wang2025ma}
Ruida Wang, Rui Pan, Yuxin Li, Jipeng Zhang, Yizhen Jia, Shizhe Diao, Renjie Pi, Junjie Hu, and Tong Zhang.
\newblock Ma-lot: Model-collaboration lean-based long chain-of-thought reasoning enhances formal theorem proving.
\newblock \emph{arXiv preprint arXiv:2503.03205}, 2025{\natexlab{c}}.

\bibitem[Wang et~al.(2024{\natexlab{b}})Wang, Ma, Zhang, Ni, Chandra, Guo, Ren, Arulraj, He, Jiang, et~al.]{wang2024mmlu}
Yubo Wang, Xueguang Ma, Ge~Zhang, Yuansheng Ni, Abhranil Chandra, Shiguang Guo, Weiming Ren, Aaran Arulraj, Xuan He, Ziyan Jiang, et~al.
\newblock Mmlu-pro: A more robust and challenging multi-task language understanding benchmark.
\newblock \emph{Advances in Neural Information Processing Systems}, 37:\penalty0 95266--95290, 2024{\natexlab{b}}.

\bibitem[Xin et~al.(2024)Xin, Ren, Song, Shao, Zhao, Wang, Liu, Zhang, Lu, Du, et~al.]{xin2024deepseek}
Huajian Xin, ZZ~Ren, Junxiao Song, Zhihong Shao, Wanjia Zhao, Haocheng Wang, Bo~Liu, Liyue Zhang, Xuan Lu, Qiushi Du, et~al.
\newblock Deepseek-prover-v1. 5: Harnessing proof assistant feedback for reinforcement learning and monte-carlo tree search.
\newblock \emph{arXiv preprint arXiv:2408.08152}, 2024.

\bibitem[Xin et~al.(2025)Xin, Xi, Yang, Chen, Wu, Xiao, Sun, Zheng, and Shen]{xin2025bfs}
Ran Xin, Chenguang Xi, Jie Yang, Feng Chen, Hang Wu, Xia Xiao, Yifan Sun, Shen Zheng, and Kai Shen.
\newblock Bfs-prover: Scalable best-first tree search for llm-based automatic theorem proving.
\newblock \emph{arXiv preprint arXiv:2502.03438}, 2025.

\bibitem[Xu et~al.(2025)Xu, Xu, Xiao, Chen, Yan, Zhang, Diao, Yang, and Wang]{xu2025ugphysics}
Xin Xu, Qiyun Xu, Tong Xiao, Tianhao Chen, Yuchen Yan, Jiaxin Zhang, Shizhe Diao, Can Yang, and Yang Wang.
\newblock Ugphysics: A comprehensive benchmark for undergraduate physics reasoning with large language models.
\newblock \emph{arXiv preprint arXiv:2502.00334}, 2025.

\bibitem[Ying et~al.(2024)Ying, Wu, Geng, Wang, Lin, and Chen]{ying2024lean}
Huaiyuan Ying, Zijian Wu, Yihan Geng, Jiayu Wang, Dahua Lin, and Kai Chen.
\newblock Lean workbook: A large-scale lean problem set formalized from natural language math problems.
\newblock \emph{arXiv preprint arXiv:2406.03847}, 2024.

\bibitem[Young \& Freedman(2019)Young and Freedman]{young2019university}
Hugh~D. Young and Roger~A. Freedman.
\newblock \emph{University Physics with Modern Physics, Global Edition}.
\newblock Pearson Education, 15th edition, 2019.
\newblock ISBN 9781292314815.

\bibitem[Yu et~al.(2025)Yu, Wan, Cheng, Zhang, Chen, Han, Wu, Yao, Hu, Ding, et~al.]{yu2025hipho}
Fangchen Yu, Haiyuan Wan, Qianjia Cheng, Yuchen Zhang, Jiacheng Chen, Fujun Han, Yulun Wu, Junchi Yao, Ruilizhen Hu, Ning Ding, et~al.
\newblock Hipho: How far are (m) llms from humans in the latest high school physics olympiad benchmark?
\newblock \emph{arXiv preprint arXiv:2509.07894}, 2025.

\bibitem[Zhang et~al.(2025)Zhang, Dong, Wu, Huang, Jia, Fernando, Shou, Zhang, and Liu]{zhang2025physreason}
Xinyu Zhang, Yuxuan Dong, Yanrui Wu, Jiaxing Huang, Chengyou Jia, Basura Fernando, Mike~Zheng Shou, Lingling Zhang, and Jun Liu.
\newblock Physreason: A comprehensive benchmark towards physics-based reasoning, 2025.
\newblock URL \url{https://arxiv.org/abs/2502.12054}.

\bibitem[Zheng et~al.(2021)Zheng, Han, and Polu]{zheng2021minif2f}
Kunhao Zheng, Jesse~Michael Han, and Stanislas Polu.
\newblock Minif2f: a cross-system benchmark for formal olympiad-level mathematics.
\newblock \emph{arXiv preprint arXiv:2109.00110}, 2021.

\bibitem[Zheng et~al.(2025)Zheng, Cheng, Yao, Wu, He, Ding, Cheng, Hu, Bai, Zhou, et~al.]{zheng2025physicsdataset}
Shenghe Zheng, Qianjia Cheng, Junchi Yao, Mengsong Wu, Haonan He, Ning Ding, Yu~Cheng, Shuyue Hu, Lei Bai, Dongzhan Zhou, et~al.
\newblock Scaling physical reasoning with the physics dataset.
\newblock \emph{arXiv preprint arXiv:2506.00022}, 2025.

\end{thebibliography}
\bibliographystyle{iclr2026_conference}

\clearpage
\appendix\label{sec:appendix}

\section{Data Sources}\label{appendix:data}
The textbook-level questions are adapted from the concepts presented in the university textbook~\citep{young2019university} and UGPhysics~\citep{xu2025ugphysics}. The Olympiad-Easy questions are derived from the intermediate solution steps of the competition questions, while the Olympiad-Hard questions are based on the problem-solving ideas in the physics Olympiad practice book ~\citep{shu2008physics}. The UGPhysics dataset is released under the CC BY-NC-SA 4.0 license. For the textbooks published by Pearson~\citep{young2019university} and Science Press~\citep{shu2008physics}, we strictly abide by their copyrights: the questions are not verbatim copies, but are rephrased and rewritten based on their physical concepts.

\section{Implementation Detail}\label{appendix:exp_setup}
The generation configuration for the LLM roll-out of the experiment is as follows:
\begin{itemize}
    \item Top-p: 0.95
    \item Temperature: 0.8
    \item Maximum tokens per generation: 16,384
    \item Repetition penalty: 1.0
\end{itemize}
Open-source models are tested under a 4-card H20 server. The entire open-source LLM roll-out process costs about 2 days.

\clearpage
\section{LLM Prompt Templates}\label{appendix:prompts}
We provide the prompt templates used to guide LLMs in generating Lean4 proofs. 
\paragraph{With \lib:}
\begin{minted}[breaklines=True, bgcolor=gray!10]{md}
Please first learn the new library besides mathlib and usage examples before answering the question. You should refer to the new unit system.

{PhysLib}

Complete the following Lean 4 code. 
Provide your response in two parts, each enclosed in separate markdown code blocks:

```plan
# Proof Plan
- Outline the main proof steps and strategies.
- Highlight key intermediate lemmas and structures.
- Describe how to connect them to form the final proof.
```
```lean4
{Lean4_header}

/-- {NL_statement} -/
{Lean4_statement}
```
\end{minted}

\paragraph{Without \lib:}
\begin{minted}[breaklines=True, bgcolor=gray!10]{md}
Complete the following Lean 4 code. 
Provide your response in two parts, each enclosed in separate markdown code blocks:

```plan
# Proof Plan
- Outline the main proof steps and strategies.
- Highlight key intermediate lemmas and structures.
- Describe how to connect them to form the final proof.
```
```lean4
{Lean4_header}

/-- {NL_statement} -/
{Lean4_statement}
```
\end{minted}

\clearpage

\section{Experiment}
\subsection{Format Study}\label{appendix:format_study}
This appendix, based on the example shown in Figure~\ref{fig:PROBLEM_FORMAT_STUDY}, conducts a more detailed analysis of the problem format mentioned in Section\ref{exp:format} with a particular focus on the characteristics of the questions themselves.

Taking a college-level problem as an example, the upper left part of Figure~\ref{fig:PROBLEM_FORMAT_STUDY} shows a force calculation problem involving two point charges on the x-axis. This problem involves various physical quantities, including charges, distances, and forces, and explicitly uses SI units such as nC and m. The problem mainly relies on numerical calculations, and the magnitude of the force can be directly applied using Coulomb's law formula, with almost no need for multiple-step formula derivations. However, due to the possible introduction of dimensional issues when combining different physical quantities for calculation, the solution to this problem needs to maintain consistency in units, which precisely demonstrates the importance of the unit system in PhysLib in formal calculations. Such problems are typical of the "formula application type" and are suitable for verifying the model's ability in basic numerical reasoning and unit handling.

For the simple problem of the competition (Competition-Easy), the problem of the parallel plate capacitor shown in the lower left corner of Figure~\ref{fig:PROBLEM_FORMAT_STUDY} is a typical example. The core formula of the problem is \(C = q/V\), and only one or two steps of substitution are required for the calculation. The problem provides values such as charge, voltage, and plate spacing, and requires sequential substitution into the formula to complete the solution. Therefore, it involves the combination of numbers and symbols. This type of problem belongs to the "transitional derivation problem", similar to the intermediate math problems in Lean's MiniF2F, mainly testing the algebraic operation ability of the model and the ability to substitute formulas. Compared to college-level problems, it has slightly increased logical reasoning requirements.

And the competition-hard problems (Competition-Hard), such as the pulley law problem shown in the right column of Figure~\ref{fig:PROBLEM_FORMAT_STUDY}, significantly increase the difficulty. The problem involves a continuous function of tension varying with angle and requires the derivation of the final result through integration or logarithmic relationships. Besides the multi-step formula derivation, the problem also includes the relationships and function expressions among multiple physical quantities, requiring an understanding of the dependencies and combination rules between the symbols in the model. Solving such problems not only requires multi-step logical analysis but also involves function operations and symbolic reasoning, typically belonging to highly difficult symbolic reasoning problems, and testing the advanced physical reasoning ability of the model.

\begin{figure*}[t]
\centering
\adjustbox{scale=1,center}{%
\begin{minipage}{\textwidth}
\lstset{basicstyle=\ttfamily\scriptsize, lineskip=-0.5ex}
\begin{minipage}[t][1.0\textheight][t]{0.495\textwidth} 
  \vspace{0pt} 
  \begin{adjustbox}{max width=\linewidth}
    \begin{tcolorbox}[title=\textit{College}, colback=white!95!gray, colframe=black!65!gray,
      boxrule=0.25pt, arc=1mm, top=0.05mm, bottom=0.05mm, left=0.5mm, right=0.5mm]
\begin{lstlisting}
Two point charges are located on the x-axis of a coordinate system: q1 = 1.0 nC is at x = +2.0 cm, and q2 = -3.0 nC is at x = +4.0 cm. What is the total electric force exerted by q1 and q2 on a charge q3 = 5.0 nC at x = 0?
theorem Electromagnetism_3_University
    (q1 q2 q3 : Charge) (x1 x2 x3 : Length)
    (hq1 : q1 = SI.nano (1  • coulomb))
    (hq2 : q2 = SI.nano (-3 • coulomb))
    (hq3 : q3 = SI.nano (5  • coulomb))
    (hx1 : x1 = ((0.02:ℝ) • meter))
    (hx2 : x2 = ((0.04:ℝ) • meter))
    (hx3 : x3 = 0)(F  : Force)
    (hF : F = K * q3 * (q1 / (x1 - x3)^2 + q2 / (x2 - x3)^2)):
    F = ((9:ℝ)*(10:ℝ)^(-22:ℚ)/32) • newton := by
  simp [←Scalar.val_inj, hF, hq1, hq2, hq3, hx1, hx2, hx3, K, SI.nano, coulomb, meter, newton]
  norm_num
\end{lstlisting}
    \end{tcolorbox}
  \end{adjustbox}

  \vspace{0.05em}

  % Medium
  \begin{adjustbox}{max width=\linewidth}
    \begin{tcolorbox}[title=\textit{Competition-Easy}, colback=white!95!gray, colframe=black!65!gray,
      boxrule=0.25pt, arc=1mm, top=0.05mm, bottom=0.05mm, left=0.5mm, right=0.5mm]
\begin{lstlisting}
The plates of a parallel-plate capacitor are 2.50 mm apart, and each carries a charge of magnitude 80.0 nC. The plates are in vacuum. The electric field between the plates has a magnitude of \(4.00 \times 10^{6}\,\text{V/m}\). What is the capacitance?
theorem Ch13_electro_question_8
    (V:Voltage)(C:Scalar capacitance_unit)(q:Charge)
    (E:Scalar (force_unit-charge_unit))(d:Length)
    (h: capacitance_unit=charge_unit - voltage_unit)
    (hC:C=(q/V).cast h)(hq:q=SI.nano (80 • coulomb))
    (hV:V=E*d)(hE:E=(4e6:ℝ) • StandardUnit _)
    (hd:d=SI.milli ((2.5:ℝ)• meter)):
    C=(1 / 125000000000:ℚ) • StandardUnit _ := by
  have hC_expanded : C = (q/(E*d)).cast h := by rw [hC, hV]
  rw [hC_expanded, hq, hE, hd]
  simp [nano, milli, coulomb, meter, ←Scalar.val_inj]
  norm_num
\end{lstlisting}
    \end{tcolorbox}
  \end{adjustbox}
  \vfill 
\end{minipage}
\hspace{0.001\textwidth} 
\begin{minipage}[t][1.0\textheight][t]{0.495\textwidth} 
  \vspace{0pt} 
  \begin{adjustbox}{max width=\linewidth}
    \begin{tcolorbox}[title=\textit{Competition-Hard}, colback=white!95!gray, colframe=black!65!gray,
      boxrule=0.25pt, arc=1mm, top=0.05mm, bottom=0.05mm, left=0.5mm, right=0.5mm]
\begin{lstlisting}
Capstan law: If a rope of coefficient of friction μ wraps n turns (θ_total = 2πn) around a post, the tension ratio between the heavy side M and the light side m satisfies n = (1 / (2πμ)) * log(M / m) assuming M > m > 0 and μ > 0.

theorem Ch2_Q1
  (M m : Mass)
  (μ : ℝ)
  (n : ℝ)
  (θ_total : ℝ := 2 * Real.pi * n)
  (T : ℝ → Force)
  (h_pos : 0 < M.val ∧ 0 < m.val ∧ 0 < μ)
  (hM_gt_m : M.val > m.val)
  (T_light_def : T 0 = m * g)
  (T_heavy_def : T θ_total = M * g)
  (capstan_differential : ∀ θ : ℝ, deriv (fun θ' => (T θ').val) θ = μ * (T θ).val)
  (capstan_integral : Real.log ((T θ_total).val / (T 0).val) = μ * θ_total)
  (theta_def : θ_total = 2 * Real.pi * n) :
  n = (1 / (2 * Real.pi * μ)) * Real.log (M.val / m.val) := by
  rcases h_pos with ⟨hM, hm, hmu⟩
  have h1 : Real.log ((T θ_total).val / (T 0).val) =
            Real.log ((M * g).val / (m * g).val) := by rw [T_heavy_def, T_light_def]
  have h2 : Real.log ((M * g).val / (m * g).val) = Real.log (M.val / m.val) := by
    have h3 : (M * g).val / (m * g).val = M.val / m.val := by
      field_simp; ring_nf; simp; ring
    rw [h3]
  have h3 : Real.log (M.val / m.val) = μ * θ_total := by linarith [capstan_integral, h1, h2]
  have h4 : Real.log (M.val / m.val) = μ * (2 * Real.pi * n) := by rw [theta_def] at h3; linarith
  have h5 : μ ≠ 0 := by linarith
  have h6 : Real.pi ≠ 0 := Real.pi_ne_zero
  have h7 : Real.log (M.val / m.val) = 2 * Real.pi * μ * n := by linarith [h4]
  have h8 : n = (Real.log (M.val / m.val)) / (2 * Real.pi * μ) := by field_simp; linarith
  rw [h8]; field_simp; ring_nf; field_simp; ring
\end{lstlisting}
    \end{tcolorbox}
  \end{adjustbox}
\end{minipage}
\end{minipage} 
}
\vspace{-23em}
\caption{Three sampled physics questions from \textit{College Textbook}, \textit{Olympics-Easy}, \textit{Olympics-Hard} problems. Each example shows the natural language problem statement followed by its corresponding Lean formalization with a verified proof.}
\label{fig:PROBLEM_FORMAT_STUDY}
\end{figure*}

\clearpage
\subsection{Case Study}\label{appendix:case_study}
\paragraph{Behavior of the same theorem with and without \lib} Figure~\ref{fig:case_differentproofs} illustrates the different proof strategies employed by Gemini-2.5-pro when solving the same college-level mechanics problem, with and without \lib. Without \lib, the proof is relatively concise and almost entirely relies on the assumptions \(ha\) and \(hT\). For the acceleration \(a\), the proof is completed by substituting the assumption into \texttt{rw [ha]} and then using basic simplification and multiplication/division commutative laws (\texttt{simp} and \texttt{mul\_div\_right\_comm}) to transform the algebraic expression. This indicates that in the absence of domain-specific libraries, LLM relies on the general tactics of Mathlib to generate proofs, which are rather mechanical but can still accomplish the task correctly. When \lib is introduced, the proof becomes more structured and systematic. After substituting the assumption \(a\), the proof utilizes the \texttt{Scalar.val\_inj} lemma from PhysLib to transform the goal into an equation in the underlying real numbers, enabling LLM to handle algebraic operations of physical quantities more directly. Subsequently, the proof is completed by applying simplification (\texttt{simp}) and \texttt{ring} tactics to prove the algebraic identity. Overall, the proof with \lib demonstrates a richer use of tools and a more robust reasoning path.

\begin{figure}[t]
\centering
\begin{adjustbox}{max width=0.99\linewidth}
\begin{tcolorbox}[title=Gemini-2.5-pro without \lib, colback=white!95!gray, colframe=black!65!gray, boxrule=0.5pt, arc=1mm, top=0mm, bottom=0mm, left=2mm, right=2mm]
\begin{lstlisting}
theorem Mechanics_73_University
  (T : Force)(m_1 m_2 : Mass)(a : Acceleration)
  (ha : a = m_2 * g / (m_1 + m_2))
  (hT : T = (m_1 * m_2) / (m_1 + m_2) * g):
  (a = (m_2 / (m_1 + m_2)) * g ∧ T = (m_1 * m_2) / (m_1 + m_2) * g) := by
  constructor
  · rw [ha]
    ext
    simp
    rw [mul_div_right_comm]
  · exact hT
\end{lstlisting}
\end{tcolorbox}
\end{adjustbox}

\vspace{0.5em}

\begin{adjustbox}{max width=0.99\linewidth}
\begin{tcolorbox}[title=Gemini-2.5-pro with \lib, colback=white!95!gray, colframe=black!65!gray, boxrule=0.5pt, arc=1mm, top=0mm, bottom=0mm, left=2mm, right=2mm]
\begin{lstlisting}
theorem Mechanics_73_University
  (T : Force)(m_1 m_2 : Mass)(a : Acceleration)
  (ha : a = m_2 * g / (m_1 + m_2))
  (hT : T = (m_1 * m_2) / (m_1 + m_2) * g):
  (a = (m_2 / (m_1 + m_2)) * g ∧ T = (m_1 * m_2) / (m_1 + m_2) * g) := by
  constructor
  . -- The first goal is to prove the rearranged expression for acceleration `a`.
    -- We start by substituting `a` with its given definition from hypothesis `ha`.
    rw [ha]
    -- The goal is now to prove an algebraic equivalence between two `Scalar` expressions.
    -- We use the `Scalar.val_inj` lemma to transform this into an equality of their
    -- underlying real values. This is a standard technique in the PhysLib library.
    simp only [← Scalar.val_inj]
    -- Now, with the goal being an equality of real numbers, we can use `simp` to apply
    -- the library's rules for how `.val` distributes over arithmetic operations.
    simp
    -- The resulting goal is `m_2.val * g.val / (m_1.val + m_2.val) = m_2.val / (m_1.val + m_2.val) * g.val`.
    -- This is a standard algebraic identity which the `ring` tactic can solve.
    ring
  . -- The second goal is to prove the expression for tension `T`, which is identical
    -- to the hypothesis `hT`.
    exact hT
\end{lstlisting}
\end{tcolorbox}
\end{adjustbox}

\caption{Two examples from \bench demonstrating different proofs of the same college-level problem, generated by Gemini-2.5-pro with and without PhysLib.}
\label{fig:case_differentproofs}
\end{figure}

\paragraph{Transfer tricks in mathematics} During the process of solving the problem depicted in Figure~\ref{fig:goedel-only-1} and Figure~\ref{fig:goedel-only}, Goedel-Prover demonstrated a variety of systematic proof strategies, which originated from its accumulated experience in mathematical Lean problems and were successfully transferred to formal proofs in physics. Firstly, the Prover would break down the overall proof goal into multiple sub-goals, such as proving $\mu_s = 0.46$ first, then proving $\mu_k = 0.40$, and finally combining them into a logical conjunction $(\mu_s = 0.46) \wedge (\mu_k = 0.40)$. This sub-goal decomposition strategy enables the Prover to handle complex problems step by step, with each sub-goal being independently verifiable and provable, thereby enhancing the reliability and success rate of automated proofs. This strategy is directly borrowed from the experience of handling complex theorems in mathematical Lean.

Within each sub-goal, the Prover will conduct a step-by-step verification of intermediate arithmetic or type values. For instance, it will first verify that \texttt{Scalar.val f\_s\_max = 230}, then \texttt{Scalar.val n = 500}, and subsequently calculate $(\mu_s : \mathbb{R}) = 0.46$ based on these intermediate results. This step-by-step verification method ensures that each stage is correct and error-free, preventing error accumulation and forming a reliable proof chain, which is exactly the same as the step-by-step verification of sub-conclusions in the Lean mathematical problems. When dealing with auxiliary steps such as type conversion or arithmetic simplification, Prover employs a multiple attempts strategy, such as using \texttt{try field\_simp at *} or \texttt{try norm\_cast at *}. These lightweight strategies enable Prover to automatically handle potential obstacles without requiring in-depth domain knowledge. The effectiveness of this strategy also stems from the experience of handling auxiliary steps through different tactics in Lean problems, achieving cross-domain transfer.

Finally, after completing the proofs of each sub-goal, the Prover uses the logical constructor to integrate the results, such as \texttt{exact ⟨h4, h7⟩}, combining the proofs of each sub-goal into an overall theorem. This method is a typical approach in mathematical Lean proofs and is also applicable in physical formalization problems, demonstrating the universality of the proof strategy and its ability for cross-disciplinary transfer.

\paragraph{Why the general model performs better} We analyzed the proof style depicted in Figure~\ref{fig:case_proofstyle} to explain why general-purpose models (such as Gemini-2.5-pro) can outperform expert theorem provers in under-trained domains. The theorem \texttt{Ch10\_question\_4} discussed involves calculating the thermodynamic index $k$ based on the known pressures $P_1, P_2$, volumes $V_1, V_2$, and temperatures $T_1, T_2$. Although both DeepSeek-Prover-V2 and Gemini-2.5-pro can complete the proof, their proof methods differ significantly. DeepSeek-Prover-V2 generates a highly step-by-step proof, gradually substituting and simplifying, explicitly introducing auxiliary steps such as $h_1$ and $h_2$, and repeatedly invoking simplification strategies for intermediate calculations; while Gemini-2.5-pro generates a concise proof by directly using all the assumptions at once and leveraging advanced simplification strategies to directly reduce the target to a numerical equation, omitting the lengthy intermediate steps.

From the perspective of the complexity of the steps, the proof of DeepSeek-Prover-V2 involves over twenty nested operations, while Gemini-2.5-pro can complete the proof through only two main strategies. This indicates that although expert proofers can ensure the rigor of the form, in this type of domain that mainly relies on numerical calculations and has less symbolic reasoning, their step-by-step operations appear cumbersome and lengthy. Further analysis reveals that DeepSeek-Prover-V2 adopts a fine-grained approach, explicitly handling each substitution and simplification, while Gemini-2.5-pro conducts reasoning at a higher abstract level and uses the built-in simplification rules in the system to quickly complete the proof. In thermodynamic problems, higher-level abstraction can reduce unnecessary complex calculations and improve efficiency.

Furthermore, the expert proofer mainly focuses on pure mathematical theorems during training. Therefore, in application fields such as thermodynamics, it is prone to introduce additional intermediate steps. This ``overthinking'' ensures rigor but may result in lengthy proofs and even increase the risk of errors in complex substitutions. In contrast, general large models possess flexible reasoning capabilities and can efficiently combine simplification rules with numerical calculations. In this case, Gemini-2.5-pro successfully mapped all hypotheses to numerical verification through a single \texttt{simp} call, demonstrating its adaptability to training deficiencies.

\begin{figure}[t]
\centering
% 上方解法：Deepseek prover v2
\begin{adjustbox}{max width=0.99\linewidth}
\begin{tcolorbox}[title=DeepSeek-Prover-V2, colback=white!95!gray, colframe=black!65!gray, boxrule=0.5pt, arc=1mm, top=0mm, bottom=0mm, left=2mm, right=2mm]
\begin{lstlisting}
theorem Ch10_question_4
    (P1 P2:Pressure) (V1 V2:Volume) (T1 T2:Temperature) (k:ℝ)
    (hV1:V1= 13• StandardUnit _)
    (hT1:T1=((1.8*15+273.15):ℝ) • kelvin)
    (hT2:T2=((-44.5*1.8+273.15):ℝ) • kelvin)
    (hP1:P1=((1.01*10^5):ℝ) • pascal)
    (hP2:P2= 868 • pascal)
    (hV2:V2=V1*T2*P1/(T1*P2))
    (hk:k=Real.rpow (V2/V1).val ((1/3):ℝ)):
    k=Real.rpow (10832250 / 144739) (1 / 3) := by
  have h_main : k = Real.rpow (10832250 / 144739) (1 / 3) := by
    have h₁ : k = Real.rpow (V2 / V1).val ((1 / 3 : ℝ)) := by rw [hk]
    rw [h₁]
    have h₂ : (V2 / V1).val = (10832250 / 144739 : ℝ) := by
      -- Simplify the expression for V2 / V1
      simp_all [Pressure, Volume, Temperature, add_assoc, add_comm, add_left_comm, mul_assoc, mul_comm, mul_left_comm, div_eq_mul_inv, mul_inv_cancel_left]
      <;> ring_nf at *
      <;> norm_num at *
      <;> field_simp [StandardUnit, mul_assoc, mul_comm, mul_left_comm] at *
      <;> ring_nf at *
      <;> norm_num at *
      <;> nlinarith
    rw [h₂]
    <;> simp [Real.rpow_def_of_pos]
    <;> norm_num
    <;> ring_nf
    <;> norm_num
    <;> field_simp
    <;> ring_nf
    <;> norm_num
    <;> nlinarith
  exact h_main
\end{lstlisting}
\end{tcolorbox}
\end{adjustbox}

\vspace{0.1em}

\begin{adjustbox}{max width=0.99\linewidth}
\begin{tcolorbox}[title=Gemini-2.5-pro, colback=white!95!gray, colframe=black!65!gray, boxrule=0.5pt, arc=1mm, top=0mm, bottom=0mm, left=2mm, right=2mm]
\begin{lstlisting}
theorem Ch10_question_4
    (P1 P2:Pressure)(V1 V2:Volume)(T1 T2:Temperature)(k:ℝ)
    (hV1:V1= 13• StandardUnit _)
    (hT1:T1=((1.8*15+273.15):ℝ) • kelvin)
    (hT2:T2=((-44.5*1.8+273.15):ℝ) • kelvin)
    (hP1:P1=((1.01*10^5):ℝ) • pascal)
    (hP2:P2= 868 • pascal)
    (hV2:V2=V1*T2*P1/(T1*P2))
    (hk:k=Real.rpow (V2/V1).val ((1/3):ℝ)):
    k=Real.rpow (10832250 / 144739) (1 / 3) := by
  -- Use all hypotheses to substitute the variables and simplify the expression.
  -- `simp` will use `hk` to substitute `k`, `hV2` to substitute `V2`,
  -- and the other hypotheses to substitute the numerical values.
  -- The `@[simp]` lemmas in the UnitsSystem library will handle the `.val` operations.
  simp [hk, hV2, hV1, hT1, hT2, hP1, hP2]
  -- The goal is now a purely numerical equality. Use `norm_num` to verify it.
  norm_num
\end{lstlisting}
\end{tcolorbox}
\end{adjustbox}
\vspace{-0.02em}
\caption{Comparison of Proof Styles by DeepSeek-Prover-V2 and Gemini-2.5-pro.}
\label{fig:case_proofstyle}
\end{figure}

\begin{figure}[t]
\centering
\begin{adjustbox}{max width=0.99\linewidth}
\begin{tcolorbox}[title=Goedel-Prover with \lib (nature statement), colback=white!95!gray, colframe=black!65!gray, boxrule=0.5pt, arc=1mm, top=0mm, bottom=0mm, left=2mm, right=2mm]
Question 74: You want to move a 500 N crate across a level floor. To start the crate moving, you have to pull with a 230 N horizontal force. Once the crate starts to move, you can keep it moving at constant velocity with only 200 N. What are the coefficients of static and kinetic friction?
Follow the process: 
Just before the crate starts to move, we have:
\[ \sum F_x = T + \big(- (f_s)_{\text{max}}\big) = 0 \quad \text{so} \quad (f_s)_{\text{max}} = T = 230~\mathrm{N} \]
\[ \sum F_y = n + (-w) = 0 \quad \text{so} \quad n = w = 500~\mathrm{N} \]
Now we solve \((f_s)_{\text{max}} = \mu_s n\), for the value of \(\mu_s\):
\[ \mu_s = \frac{(f_s)_{\text{max}}}{n} = \frac{230~\mathrm{N}}{500~\mathrm{N}} = 0.46 \]
After the crate starts to move, we have:
\[ \sum F_x = T + \big(- f_k\big) = 0 \quad \text{so} \quad f_k = T = 200~\mathrm{N} \]
\[ \sum F_y = n + (-w) = 0 \quad \text{so} \quad n = w = 500~\mathrm{N} \]
Using \(f_k = \mu_k n\), we find:
\[ \mu_k = \frac{f_k}{n} = \frac{200~\mathrm{N}}{500~\mathrm{N}} = 0.40 \]
show that the coefficients of static is $\mu_s = 0.4$ and kinetic friction is $\mu_k = 0.40$.  
\begin{lstlisting}
\end{lstlisting}
\end{tcolorbox}
\end{adjustbox}
\caption{Natural Language Statement of a Problem Solved Exclusively by Goedel-Prover.}
\label{fig:goedel-only-1}
\end{figure}

\begin{figure}[t]
\centering
\begin{adjustbox}{max width=0.99\linewidth}
\begin{tcolorbox}[title=Goedel-Prover with \lib (Lean4 statement), colback=white!95!gray, colframe=black!65!gray, boxrule=0.5pt, arc=1mm, top=0mm, bottom=0mm, left=2mm, right=2mm]
\begin{lstlisting}
theorem Mechanics_74_University_0
  (f_s_max f_k n w : Force)(μ_s μ_k : ℚ)
  (hw : w = 500 • newton)
  (hn : n = w)
  (hf_s_max : f_s_max = 230 • newton)
  (hf_k : f_k = 200 • newton)
  (hμ_s : μ_s = Scalar.val f_s_max / Scalar.val n)
  (hμ_k : μ_k = Scalar.val f_k / Scalar.val n):
  (μ_s = (0.46 : ℚ) ∧ μ_k = (0.40 : ℚ)) := by
  have h1 : Scalar.val f_s_max = 230 := by
    rw [hf_s_max]
    simp [Scalar.val_smul, newton]

  have h2 : Scalar.val n = 500 := by
    have h2₁ : n = 500 • newton := by
      rw [hn, hw]
    rw [h2₁]
    simp [Scalar.val_smul, newton]

  have h3 : (μ_s : ℝ) = (0.46 : ℝ) := by
    have h3₁ : (μ_s : ℝ) = Scalar.val f_s_max / Scalar.val n := by
      norm_cast at hμ_s ⊢
    rw [h3₁]
    rw [h1, h2]
    norm_num

  have h4 : μ_s = (0.46 : ℚ) := by
    have h4₁ : (μ_s : ℝ) = (0.46 : ℝ) := h3
    norm_num at h4₁ ⊢
    norm_num
    (try field_simp at *)
    (try norm_cast at *)

  have h5 : Scalar.val f_k = 200 := by
    rw [hf_k]
    simp [Scalar.val_smul, newton]

  have h6 : (μ_k : ℝ) = (0.40 : ℝ) := by
    have h6₁ : (μ_k : ℝ) = Scalar.val f_k / Scalar.val n := by
      norm_cast at hμ_k ⊢
    rw [h6₁]
    rw [h5, h2]
    norm_num

  have h7 : μ_k = (0.40 : ℚ) := by
    have h7₁ : (μ_k : ℝ) = (0.40 : ℝ) := h6
    norm_num at h7₁ ⊢
    (try field_simp at *)
    (try norm_cast at *)

  have h8 : (μ_s = (0.46 : ℚ) ∧ μ_k = (0.40 : ℚ)) := by
    exact ⟨h4, h7⟩

  exact h8

end Mechanics
\end{lstlisting}
\end{tcolorbox}
\end{adjustbox}
\caption{Lean4 Formalization and Proof of the Problem Shown in Figure~\ref{fig:goedel-only-1} by Goedel-Prover.}
\label{fig:goedel-only}
\end{figure}

\end{document}